\documentclass[letterpaper, 10pt, conference]{ieeeconf}

\IEEEoverridecommandlockouts
\usepackage{algorithm}
\usepackage[noend]{algpseudocode}
\usepackage{amsmath}
\usepackage{amsfonts}    
\usepackage{graphicx}
\usepackage{multicol}
\usepackage{multirow}
\usepackage{subcaption}
\usepackage{todonotes}
\usepackage{units}
\usepackage[bookmarks=true]{hyperref}
\usepackage{paralist}

\usepackage{bm}
\usepackage{bbm}
\usepackage{wrapfig}
\usepackage{booktabs, lipsum}

\usepackage{newtxtext,newtxmath}
\usepackage{mleftright}
\mleftright
\algnewcommand\algorithmicforeach{\textbf{for each}}
\algdef{S}[FOR]{ForEach}[1]{\algorithmicforeach\ #1\ \algorithmicdo}

\floatstyle{ruled}
\newfloat{algorithm}{tbp}{loa}
\providecommand{\algorithmname}{Algorithm}
\floatname{algorithm}{\protect\algorithmname}

\graphicspath{{img/}}

\title{\LARGE \bf Learning to View: Decision Transformers for Active Object Detection}
 
\author{%
  \authorblockN{%
   Wenhao Ding$^1*,2$, Nathalie Majcherczyk$^2$, Mohit Deshpande$^2$, Xuewei Qi$^2$, \\
   Ding Zhao$^1$, Rajasimman Madhivanan$^2$, Arnie Sen$^2$}
  \authorblockA{$^1$Carnegie Mellon University, $^2$Amazon Lab126\\
  }
}

\author{Wenhao Ding$^{1^*,2}$, Nathalie Majcherczyk$^1$, Mohit Deshpande$^1$, Xuewei Qi$^1$, \\
   Ding Zhao$^2$, Rajasimman Madhivanan$^1$, Arnie Sen$^1$
\thanks{$^*$Work done during the internship at Amazon.
$^1$Amazon Lab126, Sunnyvale, CA 94098, USA.
$^2$Carnegie Mellon University, Pittsburgh, PA 15213, USA.
Email: \{wenhaod, dingzhao\}@andrew.cmu.edu, \{majcherc, deshmohi, qixuewei, rajasimm, senarnie\}@amazon.com
}%
}

\begin{document}

\maketitle

\begin{abstract}

Active perception describes a broad class of techniques that couple planning and perception systems to move the robot in a way to give the robot more information about the environment.
In most robotic systems, perception is typically independent of motion planning. For example, traditional object detection is passive: it operates only on the images it receives. However, we have a chance to improve the results if we allow planning to consume detection signals and move the robot to collect views that maximize the quality of the results. 
In this paper, we use reinforcement learning (RL) methods to control the robot in order to obtain images that maximize the detection quality.
Specifically, we propose using a Decision Transformer with online fine-tuning, which first optimizes the policy with a pre-collected expert dataset and then improves the learned policy by exploring better solutions in the environment.
We evaluate the performance of proposed method on an interactive dataset collected from an indoor scenario simulator. 
Experimental results demonstrate that our method outperforms all baselines, including expert policy and pure offline RL methods.
We also provide exhaustive analyses of the reward distribution and observation space.

\end{abstract}

\section{INTRODUCTION}

\begin{figure}
  \centering
  \includegraphics[width=0.45\textwidth]{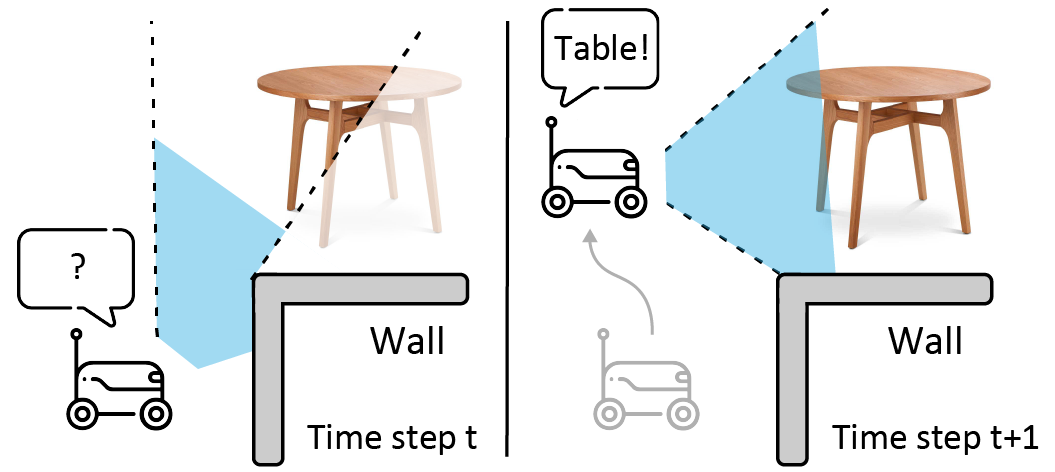}
  \caption{An example of active object detection. At time step $t$, the robot cannot detect the object. After actively moving to a new position, it can accurately detect the table.}
  \label{fig:introduction}
\end{figure}

In recent years, robot perception has improved considerably through advances in Deep Learning (DL)~\cite{lecun2015deep}. Static datasets, such as collections of Internet images~\cite{deng2009imagenet}, enabled the training of large and complex DL models~\cite{vaswani2017attention}. Traditionally, these models take in images with ad-hoc views of objects and output corresponding predictions. However, this is a passive approach to robot perception and it fails to leverage the mobility of robots. The active approach to robot perception closes the loop between robot motion and perception; it involves moving the robot to acquire sensor inputs that the perception model performs well on~\cite{zeng2020view}. In this paper, we specifically focus on the task of active object detection, which is illustrated in Figure~\ref{fig:introduction}.

%


To tackle the problem of active object detection, recent works~\cite{ammirato2017dataset, jayaraman2018end} combine object detection models with Reinforcement Learning (RL) into a decision-making framework that discovers good views by moving around a selected object~\cite{ding2020learning}.
In these settings, the RL agent maximizes a long-horizon objective by either learning from scratch in interactive environments or learning from pre-collected datasets. The latter, offline RL~\cite{levine2020offline}, is particularly suitable for robotic tasks since expert datasets can be easily obtained from demonstration or rule-based policies~\cite{fu2020d4rl}.


The major limitation of existing work comes from the Markov assumption used in traditional RL, which makes robots forget historical information that is important for memorizing visited positions. 

Language models, such as recurrent neural networks and transformers, have shown great success in the natural language processing (NLP) community because of their ability to reason about \emph{sequences} of words rather than traditional unigram or bigram models which only reason about one or two words at a time.
Decision Transformers (DT)~\cite{chen2021decision} are sequential generative models that apply this notion of reasoning about sequences of words to reasoning about tuples of (state, action, rewards) in an RL context. Although many problems in RL can be formulated as a Markov Decision Process (MDP), DT shows that breaking the Markov assumption to reason about sequences of tuples outperforms traditional methods.

For active object detection, previous work relies heavily on RGB images which do not provide any spatial navigation information or information about the obstacle configuration~\cite{yang2019embodied}. Incorporating some depth-based representation, such as a depth image, into the state would provide this missing information to the model to avoid obstacles and occlusions. Additionally, previous work suffers from training instability since they attempt to jointly optimize for the detector and the motion policy~\cite{kotar2022interactron}.

Furthermore, models trained using only offline RL methods do not generalize to out-of-distribution states and have difficulties reaching the optimal policy.
In this paper, we propose to combine online and offline training on a motion policy that traverses favorable views. We first leverage expert data to pre-train the policy model in an offline manner and then fine-tune it using an interactive synthetic dataset. This type of dataset consists of a set of views collected from robot poses throughout the environment. 
We also propose a new sampling procedure to increase the exploration of DT for better online fine-tuning.
To evaluate our proposed method, we collect an interactive dataset from AI2THOR~\cite{kolve2017ai2} and use it for online fine-tuning and expert dataset collection.
Experiment results demonstrate that our method dramatically outperforms the expert policy.

Our contributions can be summarized as follows:
\begin{itemize}
    \item We build upon existing online DT work~\cite{zheng2022online} to tackle the active object detection problem. Our approach combines offline pre-training and online fine-tuning of a DT for active vision. We use a novel sampling method in the training of the algorithm to increase RL exploration.
    \item We collect an interactive dataset in the AI2THOR simulator based on the protocol and detection model from~\cite{kotar2022interactron}, which contains numerous scenes for the active object detection task.
    \item We provide exhaustive analyses of our proposed method and a thorough comparison with strong baselines.
\end{itemize}

We organize this paper as follows. We first discuss related work in Section~\ref{sec:relatedwork}. Then, we formalize the problem statement in Section~\ref{sec:problem}. The design of our method is presented in Section~\ref{sec:method}. We report the results of our performance evaluation in Section~\ref{sec:experiment}, and conclude the paper in Section~\ref{sec:conclusion}.

\section{RELATED WORK}
\label{sec:relatedwork}

\subsection{Active Object Detection}

Embodied AI~\cite{duan2022survey} is an emerging topic that learns through interactions with the environment from an egocentric perception instead of learning from static datasets of images, videos, or text from the internet. Active object detection is one typical task of embodied AI and has been studied for many years.
Ammirato et al.~\cite{ammirato2017dataset} propose a benchmark for active vision tasks that includes an interactive dataset. They implement a baseline approach that uses the REINFORCE~\cite{williams1992simple} algorithm with the object detection score as the reward. 
\cite{jayaraman2018end} also relies on RL but adds an aggregator module to accumulate historical information to find the next best view.
\cite{yang2019embodied} uses an occlusion ratio as the reward to train the motion policy and proposes a three-stage pipeline. They first train the detection model, then train the motion policy, and finally fine-tune the detection model to achieve better performance.
\cite{fang2020move} collects multi-view data, generates pseudo-labels, and fine-tunes the object detector in a self-supervised manner.

Besides the RL formulation, a meta-learning framework~\cite{hospedales2021meta} is also used to solve the active object detection task~\cite{kotar2022interactron, wortsman2019learning}. Instead of maximizing a reward that indicates the performance, they train the detection model on a sequence of images to maximize the detection results of the first frame. Thus the policy model aims to find images that can help the detection model quickly adapt to the first frame.

\subsection{Decision Transformer}

Transformers~\cite{vaswani2017attention} achieve great success in NLP tasks, such as language translation and sentence prediction. They use the self-attention mechanism, which computes the correlation between the inputs and outputs.

Recently, researchers adapt Transformer models to decision-making tasks~\cite{chen2021decision, janner2021offline} since it can be used for action prediction conditioned on historical states and reward-to-go. Although trained in a supervised learning manner, DT outputs different actions to match different target reward-to-go.
More variants of DT are proposed later, for example, \cite{paster2022you} extends DT to stochastic environments, \cite{boustati2021transfer} incorporates counterfactual reasoning into DT, and \cite{furuta2021generalized} generalizes DT to broader usage for hindsight information matching.

\subsection{Offline Reinforcement Learning}

RL algorithms~\cite{haarnoja2018soft, schulman2017proximal, ding2022generalizing} learn from the interaction between agent and environment, usually leading to inefficient training and dangerous exploration. 
Although off-policy~\cite{haarnoja2018soft} algorithms make it possible to use a replay buffer for model training, they build the buffer with data collected from their online learning policy. 
Recently, some ``truly'' off-policy algorithms, also known as offline RL, build the buffer with data collected from human experts or rule-based methods~\cite{prudencio2022survey}. 

Offline RL can be simply implemented~\cite{kumar2020conservative} by adding regularization to DQNs~\cite{mnih2013playing}, which avoids the over-estimation of unseen action-state value in the pre-collected dataset. 
To avoid evaluating unseen actions outside of the dataset, \cite{kostrikov2021offline} proposes IQL that implicitly approximates the policy improvement step.
More advanced variants of offline DQN are introduced in~\cite{agarwal2020optimistic}.

Most offline RL methods are limited by the quality of the dataset, e.g., a dataset collected from a sub-optimal policy makes the agent never see optimal states. 
Two solutions are proposed to break the limitation of the low-quality dataset. One is encouraging exploration during dataset generation~\cite{yarats2022don} and the other is adding an online fine-tune stage after offline training~\cite{nair2020awac, zheng2022online}.

\section{PROBLEM STATEMENT}
\label{sec:problem}

\begin{figure*}
  \centering
  \includegraphics[width=0.9\textwidth]{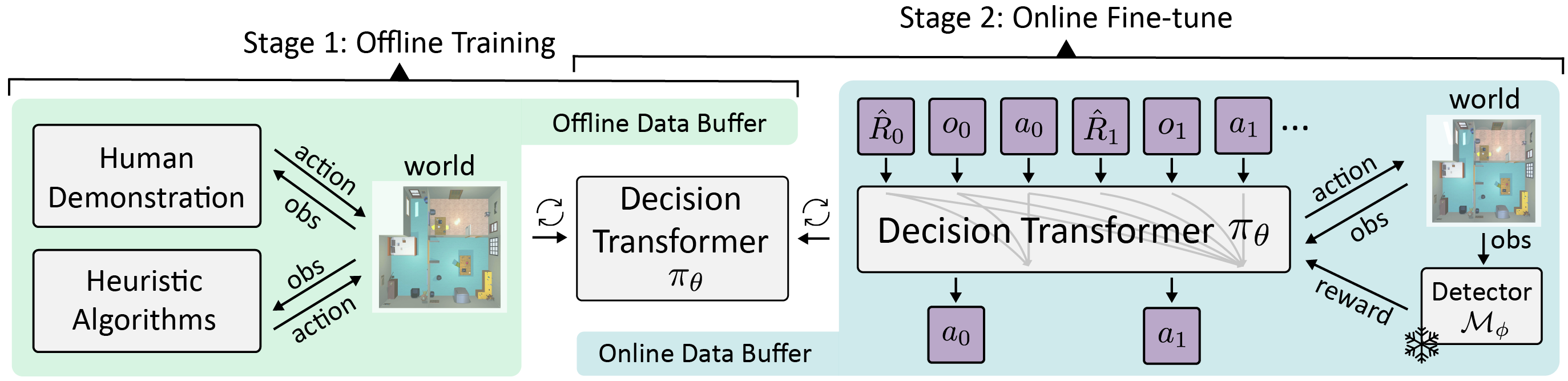}
  \caption{Illustration of the proposed framework. In Stage 1, we conduct offline training on the Decision Transformer with expert data collected from human or heuristic algorithms. In Stage 2, we fine-tune the model in an interactive environment, where the reward comes from a detection model with frozen parameters.}
  \label{fig:pipeline}
\end{figure*}

\subsection{Objective}
\label{sec:objective_notation}

The aim of active object detection is to learn a motion policy $\pi_{\theta}$ to control a mobile robot such that the predictions of the object detection model $\mathcal{M}_{\phi}$ are maximal.
The detection model predicts bounding boxes $\hat{b}$ of objects over RGB images such that $\mathcal{M}_{\phi}: I^{RGB}\rightarrow (\hat{b}, \delta)$, where $\delta$ is the detection confidence score associated with the prediction. $\delta$ is set to 0 if the prediction is wrong. We denote the corresponding depth image as $I^{D}$ and $T$ refers to the maximum number of steps that can be taken to improve the detection performance.

\subsection{Assumptions}

To focus on the core problem, we make several important assumptions in our framework:
\begin{itemize}
    \item To avoid training instability, we assume the weights of the detection model $\mathcal{M}_{\phi}$ are frozen. 
    \item We discretize the state and action spaces to simplify dataset collection. 
    \item We use the detection score as a surrogate value for detection performance. Although previous work~\cite{gallos2019active} claims that this score may not be a good metric, we find it still useful as shown in our experimental analysis.
    \item We assume that the robot has access to depth images, which allow us to consider environmental context and distances to surrounding objects. 
\end{itemize}

\subsection{Formulation}

We formulate this problem as a Partially Observable Markov Decision Process (POMDP), where the agent cannot directly observe the underlying state.
$s \in \mathcal{S}$ represents the internal state, including parameters $\phi$ of $\mathcal{M}_{\phi}$ and properties of the object to be detected.
$o \in \mathcal{O}(o|s', a)$ represents the observable state to the policy model, which contains the bounding box of selected object $\hat{b}$, the RGB image $I^{RGB}$, and the depth image $I^{D}$.
$a \in \mathcal{A}$ represents the output action of the policy, which is selected from the set $\mathcal{A}=\{\uparrow,\downarrow,\leftarrow,\rightarrow,\circlearrowleft,\circlearrowright, \times \}$.
Finally, the reward $r \in \mathcal{R}$ represents the performance of the detection model with $r = \delta (I^{RGB})$.
Our goal is to maximize the expected reward $\max_{\theta} \mathbb{E}_{\pi_{\theta}}[\sum_{t=0}^{T} r_t$], where the policy model $\pi_{\theta}$ takes as input an observation $o$ and outputs an action $a$.

\section{METHODOLOGY}
\label{sec:method}

In this paper, we propose to tackle the Active Object Detection problem using a DT as the policy model $\pi_{\theta}$ and introduce it in Section~\ref{sec:policy_model}. We also discuss the offline-online learning framework designed for efficiently training this model in Section~\ref{sec:offline_online}. The entire procedure of the algorithm is summarized in Algorithm~\ref{algorithm1}.

\subsection{Policy Model Design}
\label{sec:policy_model}

The key idea behind Decision Transformers is conditional action generation, where the output action is conditioned on the desired future rewards represented by the reward-to-go (RTG):
\begin{equation}
    \hat{R}_t = \sum_{t'=t}^T r_{t'},
\end{equation}
where $t$ and $T$ are the current time step and the final time step.
Decision Transformers take in as input a sequence of RTGs, observations, and actions. We refer to this sequence as trajectory $\tau$:
\begin{equation}
    \tau = (\hat{R}_0, o_0, a_0,\dots, \hat{R}_t, o_t).
\end{equation}
The ground truth of the output sequence is $a_t$. Since we use discrete actions, we select cross entropy $\mathcal{L}_{CE}$ as the training loss function.
At inference time, the true RTG is not known so we approximate it using an estimate of the maximum accumulated reward for the task.
Since the maximum reward for each time step is in $[0, 1]$ as we use the detection score, we set the initial RTG $\hat{R}_0$ to the length of episode $T$. Then, the RTG is updated by subtracting the true reward obtained at each step $\hat{R}_{t+1} = \hat{R}_{t} - r_t$. 

The training of DT follows the same protocol as supervised training and thus requires batch sampling from the data buffer. In the original DT paper~\cite{chen2021decision}, the probability of sampling one trajectory is proportional to the length of the trajectory. 
We sample the data buffer proportional to the variance of the rewards in the trajectory to encourage exploration and allow the model to see a diverse set of samples.
Since the variance of the states is difficult to calculate, we use the variance of the rewards as a tractable proxy.
Specifically, for the $i$-th trajectory, the probability is
\begin{equation}
    p_i = \frac{Var(\tau_i)}{\sum_{i}^N Var(\tau_i)}
\end{equation}
where $N$ is the total number of training trajectories.
To further increase exploration, we add an entropy term $\mathcal{L}_{en}$ of the predicted action, resulting in the final loss function:
\begin{equation}
    \mathcal{L} = \mathcal{L}_{CE} - \lambda \mathcal{L}_{en},
\label{equ:objective}
\end{equation}
where $\lambda$ is empirically set to 0.1 for balancing the weight of two training losses.

\begin{algorithm}[t]
\caption{Offline-Online Training of DT}
\label{algorithm1}
\begin{algorithmic}[1]
\State \textbf{Input:} Expert Policy $\pi_{e}$, Buffer Size $N$, Simulator $\mathcal{E}$
\State \textbf{Output:} Motion Policy $\pi_{\theta}$ 
\State  $\triangleright$  \textbf{Stage 1: Offline Training} 
\For{$i$ in [1, $N$]}
    \State Roll out $\pi_{e}$ in $\mathcal{E}$ to get trajectory $\tau_i$, $\mathcal{B} \leftarrow \mathcal{B} \cup \{\tau_i\}$ 
\EndFor \\
Train $\pi_{\theta}$ with $\mathcal{L}$ with buffer $\mathcal{B}$ 

\State $\triangleright$ \textbf{Stage 2: Online Fine-tuning}
\While{\text{not converged}}
    \State Roll out $\pi_{\theta}$ in $\mathcal{E}$ to get trajectory $\tau_i$
    \State Remove the first point in $\mathcal{B}$, $\mathcal{B} \leftarrow \mathcal{B} \cup \{\tau_i\}$ 
    \State Train $\pi_{\theta}$ with $\mathcal{L}_{CE}$ with buffer $\mathcal{B}$ 
\EndWhile
\end{algorithmic}
\end{algorithm}

\begin{figure*}
  \centering
  \includegraphics[width=0.98\textwidth]{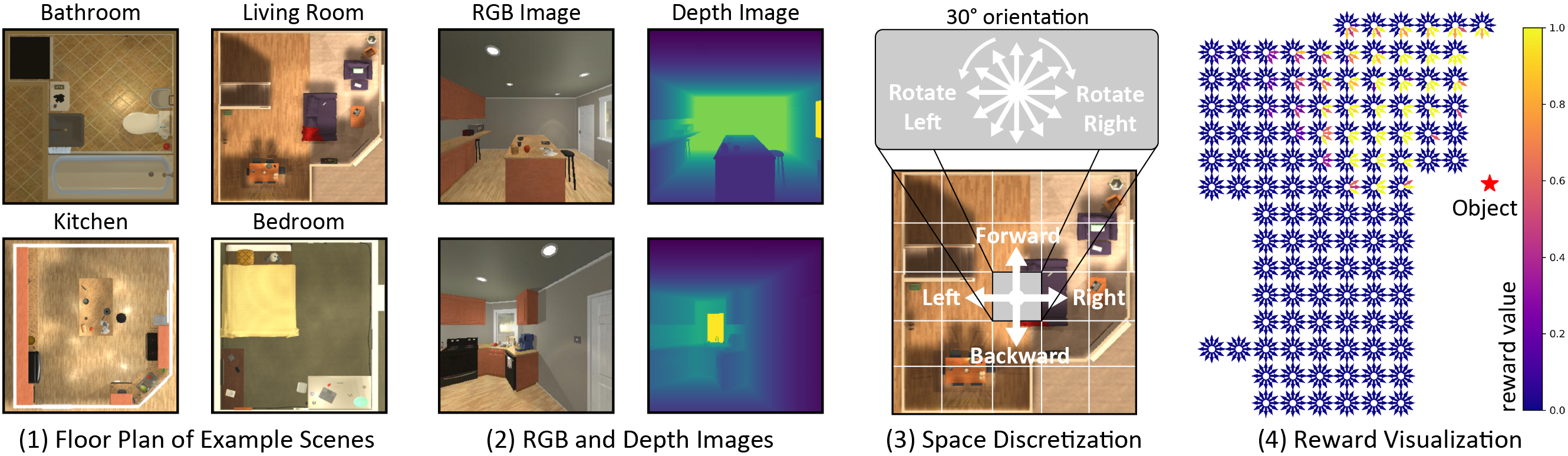}
  \caption{Description of the interactive dataset. (1) The floor plan of 4 categories of scenes provided by the AI2THOR simulator. (2) The RGB and depth images obtained from the simulator. (3) Discretization of the location and orientation together with 7 discrete actions. (4) Reward (detection score) of one object in the discretized dataset.}
  \label{fig:simulator}
\end{figure*}

\begin{figure*}
  \centering
  \includegraphics[width=1.0\textwidth]{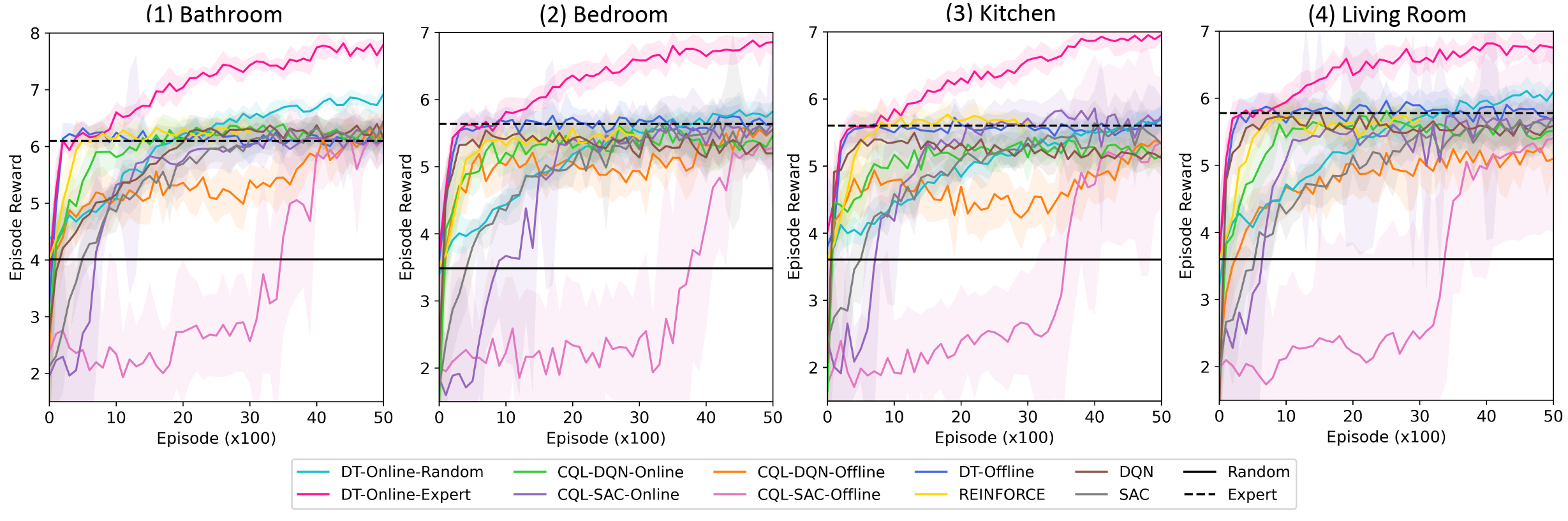}
  \caption{Episode reward of proposed method and baselines in 4 categories of floor plans.}
  \label{fig:reward}
  \vspace{-3mm}
\end{figure*}

\subsection{Offline and Online Training Framework}
\label{sec:offline_online}

We illustrate the proposed offline-online training framework in Figure~\ref{fig:pipeline}. The original DT requires a data buffer $\mathcal{B}$ that stores the pre-collected dataset. To generate such a dataset, we design a heuristic expert that imitates human logic: it first rotates to put the object in the middle of the view and then goes forward to minimize the distance to the object. Note that this simple heuristic policy is by no means optimal since the selected object could be occluded by other objects in the scene. Purely training the DT on this dataset makes it almost impossible to achieve optimal performance, thus we still need to obtain new information from the interactive environment.

Therefore, we propose to add an online training stage after the offline training, which gradually replaces samples in the data buffer $\mathcal{B}$ with new data obtained by the learned policy $\pi_{\theta}$. After $10$ episodes of online data collection, we use the new data buffer to train the policy $\pi_{\theta}$ and repeat this process until reaching the maximum number of episodes. 
When collecting the new data sample, we use the true RTG calculated from the reward rather than the approximated RTG. This procedure is also called Hindsight Return Relabeling, which is explored in a similar offline-online framework in~\cite{zheng2022online}.

\section{EXPERIMENTS}
\label{sec:experiment}

We first describe the design and collection process of our interactive dataset in Section~\ref{sec:data_collection}. Then, we introduce the overall performance of our method compared against several baselines in Section~\ref{sec:performance}. Finally, we study the reward distribution  in Section~\ref{sec:reward_distribution} and perform an ablation study in Section~\ref{sec:performance}.

\subsection{Interactive Dataset Collection}
\label{sec:data_collection}

AI2THOR~\cite{kolve2017ai2} is a popular simulator for embodied AI and is used by Kotar et al.~\cite{kotar2022interactron} to collect their interactive dataset. Since we propose a new formulation of the active object detection problem, we create a new dataset. There are 4 categories of scenes (\textit{Bathroom}, \textit{Kitchen}, \textit{Living Room}, and \textit{Bedroom}) as shown in Figure~\ref{fig:simulator} (1). Each scene contains 30-floor plans. The robot navigates in the 2D scene and acquires observations shown in Figure~\ref{fig:simulator} (2), including RGB images and depth images.

\begin{table*}[th]
\caption{Detection results of different motion policies.}
\label{tab:evaluation}
\centering
\begin{tabular}{p{2.3cm}|p{0.6cm}p{0.8cm}p{0.95cm}|p{0.6cm}p{0.8cm}p{0.95cm}|p{0.6cm}p{0.8cm}p{0.95cm}|p{0.6cm}p{0.8cm}p{0.95cm}}
\toprule
\multirow{2}{*}{Method}  & \multicolumn{3}{c|}{Bathroom} & \multicolumn{3}{c|}{Bedroom} & \multicolumn{3}{c|}{Kitchen} & \multicolumn{3}{c}{Living Room} \\
                    & Recall  & Precision  & Reward  & Recall  & Precision  & Reward  & Recall  & Precision  & Reward   & Recall  & Precision  & Reward \\
\midrule
Random              & 0.332 & 0.374 & 4.03\scriptsize{$\pm$0.10}            & 0.254  & 0.300 & 3.58\scriptsize{$\pm$0.09}            
                    & 0.246 & 0.297 & 3.59\scriptsize{$\pm$0.11}            & 0.260  & 0.325 & 3.60\scriptsize{$\pm$0.15}  \\
Expert              & 0.554 & 0.828 & 6.10\scriptsize{$\pm$0.16}            & 0.466  & 0.867 & 5.62\scriptsize{$\pm$0.19}            
                    & 0.385 & 0.755 & 5.66\scriptsize{$\pm$0.01}            & 0.500  & 0.853 & 5.59\scriptsize{$\pm$0.21}  \\
\midrule
REINFORCE           & 0.567 & 0.832 & 6.24\scriptsize{$\pm$0.21}            & 0.441  & 0.837 & 5.43\scriptsize{$\pm$0.18}            
                    & 0.376 & 0.764 & 5.66\scriptsize{$\pm$0.14}            & 0.532  & 0.869 & 5.70\scriptsize{$\pm$0.17}  \\
DQN                 & 0.570 & 0.826 & 6.15\scriptsize{$\pm$0.23}            & 0.402  & 0.815 & 5.20\scriptsize{$\pm$0.30}           
                    & 0.381 & 0.742 & 5.12\scriptsize{$\pm$0.28}            & 0.461  & 0.827 & 5.59\scriptsize{$\pm$0.26}  \\
SAC                 & 0.535 & 0.814 & 6.45\scriptsize{$\pm$0.30}            & 0.442  & 0.861 & 5.55\scriptsize{$\pm$0.47}            
                    & 0.373 & 0.731 & 5.33\scriptsize{$\pm$0.23}            & 0.487  & 0.823 & 5.73\scriptsize{$\pm$0.23}  \\
\midrule
CQL-DQN-Offline     & 0.535 & 0.824 & 6.44\scriptsize{$\pm$0.19}            & 0.456  & 0.852 & 5.56\scriptsize{$\pm$0.29}            
                    & 0.331 & 0.693 & 5.35\scriptsize{$\pm$0.31}            & 0.320  & 0.754 & 5.10\scriptsize{$\pm$0.23}  \\
CQL-SAC-Offline     & 0.563 & 0.837 & 6.17\scriptsize{$\pm$0.46}            & 0.418  & 0.824 & 5.28\scriptsize{$\pm$0.24}            
                    & 0.314 & 0.703 & 5.35\scriptsize{$\pm$0.97}            & 0.481  & 0.832 & 5.41\scriptsize{$\pm$1.40}  \\
DT-Offline          & 0.546 & 0.821 & 6.12\scriptsize{$\pm$0.29}            & 0.472 & 0.859 & 5.62\scriptsize{$\pm$0.15}            
                    & 0.392 & 0.742 & 5.74\scriptsize{$\pm$0.18}            & 0.515  & 0.848 & 5.67\scriptsize{$\pm$0.23}  \\
\midrule
CQL-DQN-Online      & 0.560 & 0.831 & 6.21\scriptsize{$\pm$0.19}            & 0.435  & 0.853 & 5.46\scriptsize{$\pm$0.32}            
                    & 0.392 & 0.748 & 5.15\scriptsize{$\pm$0.16}            & 0.463  & 0.832 & 5.51\scriptsize{$\pm$0.34}  \\
CQL-SAC-Online      & 0.552 & 0.811 & 6.11\scriptsize{$\pm$0.34}            & 0.452  & 0.871 & 5.65\scriptsize{$\pm$1.35}            
                    & 0.371 & 0.750 & 5.72\scriptsize{$\pm$0.31}            & 0.503  & 0.849 & 5.68\scriptsize{$\pm$0.37}  \\
DT-Online-Random    & 0.672 & 0.875 & 6.93\scriptsize{$\pm$0.19}            & 0.478  & 0.853 & 5.82\scriptsize{$\pm$0.14}            
                    & 0.369 & 0.743 & 5.69\scriptsize{$\pm$0.20}            & 0.510  & 0.851 & 6.10\scriptsize{$\pm$0.19}  \\
DT-Online-Expert    & \textbf{0.734} & \textbf{0.901} & \textbf{7.80\scriptsize{$\pm$0.10}}            & \textbf{0.785}  & \textbf{0.907} & \textbf{6.86\scriptsize{$\pm$0.21}}            
                    & \textbf{0.692} & \textbf{0.876} & \textbf{6.96\scriptsize{$\pm$0.08}}            & \textbf{0.779}  & \textbf{0.898} & \textbf{6.75\scriptsize{$\pm$0.27}}  \\
\bottomrule
\end{tabular}
\vspace{-3mm}
\end{table*}

The reason we collect this dataset instead of directly using the simulator is that we can avoid online rendering by querying stored images to speed up the interaction process. However, we cannot store all images for all continuous locations. Therefore, we discretize the space into 30~cm-sized cells as shown in Figure~\ref{fig:simulator} (3). For each cell, we quantize the orientation into 30~degree increments. The action space is a set of actions that move the robot linearly or angularly by these increments. 
When the robot takes an action, we query the corresponding image from the dataset. 

\begin{figure}
  \centering
  \includegraphics[width=0.49\textwidth]{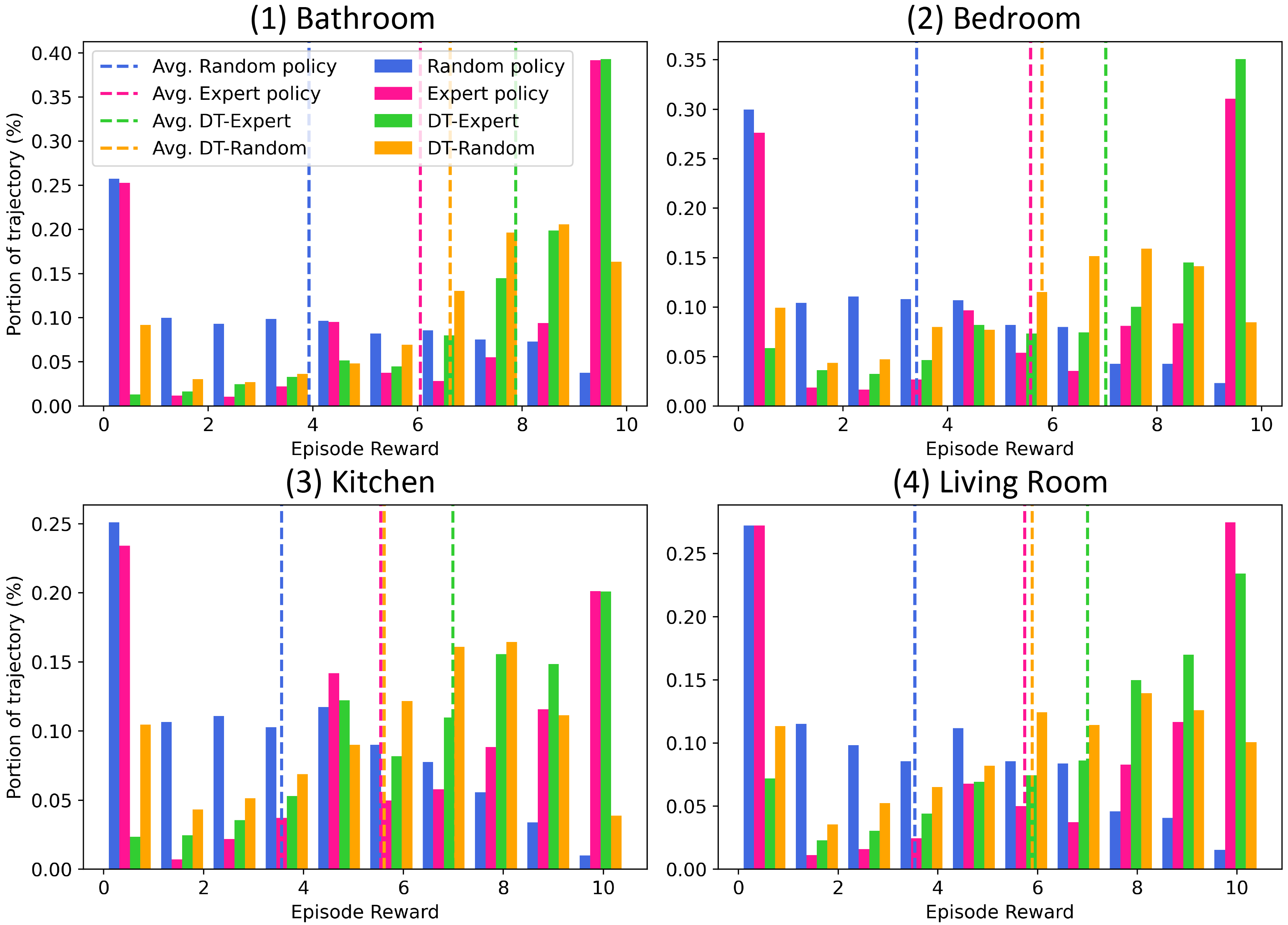}
  \caption{Distribution of episode reward on 4 scenes. The expert policy is not optimal since it fails in $\sim$25\% of cases. In contrast, our method reduces the portion of failure cases to $\sim$5\%.}
  \label{fig:scene_bar}
\end{figure}

The last missing part of the dataset is the reward. As introduced in Section~\ref{sec:objective_notation}, we use the detection score of the model $\mathcal{M}_{\phi}$ as the reward. To generate such rewards, we use the DETR~\cite{carion2020end} model with pre-trained weights provided by~\cite{kotar2022interactron}. We plot an example of the reward over the reachable space for one object in Figure~\ref{fig:simulator} (4). We observe that the value of the reward is consistent with the distance between the robot and the object, and the reward drops to 0 if the robot is too close.

Not all poses are meaningful to start from. Poses where the object is already clearly in view or poses where we do not get a detection do not start useful trajectories for training the model. We apply several rules to select the initial poses: the reward is lower than a threshold of $0.2$, the area of bounding box $\hat{b}$ is larger than 200, and the distance between the object and the robot is larger than $3$ m. After applying these rules, we obtain $\sim$1000 initial poses in each scene category. 


\begin{figure*}[ht]
  \centering
  \includegraphics[width=0.98\textwidth]{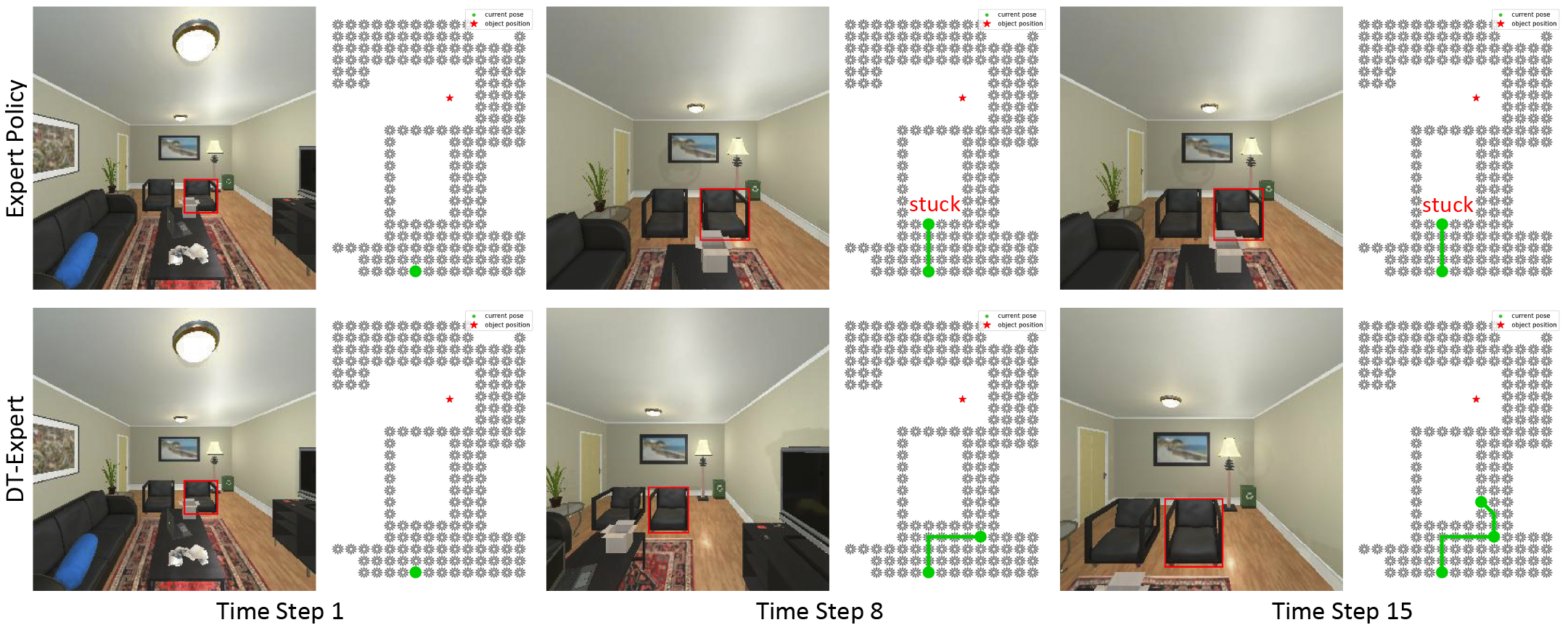}
  \caption{Challenging example with an armchair as the selected object. The top row shows the trajectory generated by the expert policy and the bottom row shows the trajectory generated by DT-Expert. The expert is stuck by the table while the DT-Expert bypasses the table and finds a good view of the armchair.}
  \vspace{-3mm}
  \label{fig:failure_cases}
\end{figure*}

\subsection{Baselines}
\label{sec:performance}
To focus on the offline and online training framework, we consider 4 types of baseline models: 
\begin{itemize}
    \item Heuristic Algorithms. The simplest rule is randomly selecting actions from the action set, which is named \textbf{Random}. We also design a human-inspired algorithm named \textbf{Expert} that first rotates to put the object in the middle of the view and then moves forward.
    \item Online RL. \textbf{DQN}~\cite{mnih2013playing} and \textbf{SAC}~\cite{haarnoja2018soft} are two well-known off-policy RL methods that can be used with a discrete action space. \textbf{REINFORCE}~\cite{williams1992simple} is used by Ammirato et al.~\cite{ammirato2017dataset} as the motion policy.
    \item Offline RL. CQL~\cite{kumar2020conservative} is a state-of-the-art offline RL method, which is implemented based on either DQN (\textbf{CQL-DQN-Offline}) or SAC (\textbf{CQL-SAC-Offline})). We also consider the original version \textbf{DT-Offline} as a baseline. All these methods are trained offline with a data buffer collected by the expert policy.
    \item Offline RL with Online Training. We adapt the DQN and SAC flavors of CQL for online training and name them \textbf{CQL-DQN-Online} and \textbf{CQL-SAC-Online}. \textbf{DT-Online-Random} is a special case that is first trained with a data buffer collected by the random policy and then fine-tuned online.
\end{itemize}
All these methods are implemented in PyTorch~\cite{paszke2017automatic}. 
The training of motion policy runs on an NVIDIA RTX 3080Ti GPU with 12 GB memory. 

\subsection{Performance Evaluation}
\label{sec:performance}

We show the episode reward results of our method (\textbf{DT-Online-Expert}) and all baselines in Figure~\ref{fig:reward}. We note that our method outperforms all other methods including the \textbf{Expert} method. Compared with DT-Online-Random, our method bootstraps from the expert policy and outperforms it through online interaction with the environment. DT-Offline only bootstraps from the expert policy without collecting additional samples from the environment. Therefore, the advantage of DT-Online-Expert comes from both offline pre-training and online fine-tuning. Lastly, although CQL-DQN-Online and CQL-SAC-Online share the same training framework as ours, CQL fails to achieve the same performance as our method. Intuitively, this indicates that the Markov assumption fails to capture the complexity of this problem and that temporal information benefits the training of the motion policy.

Since we use the detection score as a surrogate metric for the performance of the object detection model, we need to validate that this reward captures the real objective: maximizing the detection quality. We compute the precision and recall of the detector along the trajectory (Table~\ref{tab:evaluation}) and show that our method still achieves the best results, which indicates that using the detection score is reasonable and the motion policy indeed improves the detection quality.


\subsection{Reward Distribution}
\label{sec:reward_distribution}

To further analyze the performance of our approach, we look into which scenarios can be solved by our method but not by the expert. We plot the distribution of episode rewards from 4 different methods in Figure~\ref{fig:scene_bar}. The x-axis is the episode reward and the y-axis is the portion of scenarios.

Unsurprisingly, the distribution of rewards of the random policy concentrates on the low-reward region and receives a reward of 0 in $\sim$25\% of scenarios. However, we notice that the expert policy receives a reward of 10 in $\sim$25\% of scenarios but also a reward of 0 in another $\sim$25\% of scenarios, which indicates that the expert policy has a lot of failure cases. 

Now let us turn to learning-based methods. DT-Random reduces the number of low-reward scenarios but cannot achieve high rewards in most scenarios. Our DT-Expert method further reduces the number of low-reward scenarios and perfectly solves most scenarios. In summary, DT-Expert outperforms the baselines because it solves some challenging cases that cannot be solved by other methods. We show one example of the challenging cases in Figure~\ref{fig:failure_cases} for better illustration. We find that Expert fails because the robot is stuck by an obstacle in the scene. In contrast, DT-Expert bypasses this obstacle using the depth information from depth images.

\begin{table}[th]
\caption{Ablation study of observation space.}
\label{tab:ablation}
\centering
\begin{tabular}{c|cccc}
\toprule
Scenario       & w/o Both  & w/o Depth & w/o RGB & Depth+RGB \\ 
\midrule
Bathroom       & 6.10\scriptsize{$\pm$0.14} & 6.14\scriptsize{$\pm$0.17} & 7.75\scriptsize{$\pm$0.12} & 7.80\scriptsize{$\pm$0.10} \\
Bedroom        & 5.72\scriptsize{$\pm$0.17} & 5.70\scriptsize{$\pm$0.21} & 6.84\scriptsize{$\pm$0.13} & 6.86\scriptsize{$\pm$0.21} \\
Kitchen        & 5.58\scriptsize{$\pm$0.08} & 5.61\scriptsize{$\pm$0.05} & 6.95\scriptsize{$\pm$0.15} & 6.96\scriptsize{$\pm$0.08} \\
Living Room    & 5.52\scriptsize{$\pm$0.13} & 5.52\scriptsize{$\pm$0.18} & 6.75\scriptsize{$\pm$0.21} & 6.75\scriptsize{$\pm$0.27} \\
\bottomrule
\end{tabular}
\vspace{-0mm}
\end{table}

\subsection{Ablation study}
\label{sec:ablation}

To investigate the influence of the observation space on the DT model, we conduct two ablation experiments, one which removes the RGB image and one which removes both the RGB image and the depth image. We notice that the RGB image provides limited information compared to the depth image since active object detection heavily relies on the distance between the robot and the object as well as the obstacles in front of the robot.
Since the depth image provides more information than the RGB image, if we had a local map around the robot, we suspect that would provide even more spatial information to the DT model to better solve the task.


\section{CONCLUSIONS}
\label{sec:conclusion}

In this paper, we focused on the active object detection problem, where a motion policy is trained to control the robot to achieve better detection performance.
To efficiently investigate this problem, we collected an interactive dataset in the AI2THOR simulator, used the detection score of a pre-trained detection model as the reward, and designed several rules for initial pose selection.
Then, we applied Decision Transformers to this problem and proposed a novel offline-online training framework. 
We demonstrated the proposed method efficiently combines the advantages of offline expert information and online interactions with the environment. 
Further analysis of scenario difficulty and ablation studies illustrated that the spatial context is important to avoid obstacles and find the best view of the selected object.

There are still critical limitations of the proposed method. We use the detection score to indicate the detection performance, and this metric is unreliable for out-of-distribution inputs to the detection model. Finding other metrics to measure the uncertainty of detection will be an important future direction. Stably training the detection and motion models is another avenue to explore.



\clearpage

\bibliographystyle{IEEEtran}
\bibliography{references}

\end{document}